# IMPROVING THE EFFICIENCY OF A GENERATION ALGORITHM FOR SHAKE AND BAKE MACHINE TRANSLATION USING HEAD-DRIVEN PHRASE STRUCTURE GRAMMAR

**Fred Popowich**

School of Computing Science, Simon Fraser University
Burnaby, British Columbia, CANADA V5A 1S6
popowich@cs.sfu.ca

**Abstract**

A Shake and Bake machine translation algorithm for Head-Driven Phrase Structure Grammar is introduced based on the algorithm proposed by Whitelock for unification categorial grammar. The translation process is then analysed to determine where the potential sources of inefficiency reside, and some proposals are introduced which greatly improve the efficiency of the generation algorithm. Preliminary empirical results from tests involving a small grammar are presented, and suggestions for greater improvement to the algorithm are provided.

## 1.0  Introduction

Traditional approaches to Machine Translation (MT) can be classified into two general areas, those that use a *transfer* based approach and those that use an *interlingua* [5]. It is highly appealing to apply the declarative approach of unification based grammar formalisms to the specific task of machine translation (MT), which has traditionally been very procedural in nature as noted in [10].

Shake and Bake (S&B) is an approach to MT which "overcomes some difficulties encountered by transfer and interlingua methods" [1]. In an S&B MT system, unilingual lexicons of different languages are connected by a multilingual lexicon which establishes relationships (via unification) between sets of lexical entries in the various languages. One advantage of the S&B approach is that unilingual grammars and lexicons can be developed independently. Prototype systems, using Unification Categorial Grammar (UCG) [3] have been constructed for bidirectional English-Spanish translations [1] and for multi-directional English-French-Japanese translations (Whitelock, pers. comm.).

In this paper, we adapt the S&B approach taken for UCG, and apply it to Head-Driven Phrase Structure Grammar (HPSG) [8],[9], one of the most widely studied unification-based grammar formalisms. Whereas UCG relies on binary grammar rules, in HPSG there is no restriction on the branching factor associated with a grammar rule. Furthermore, HPSG relies on a collection of *principles* which incorporate intricate functional and relational dependencies between constituents. After introducing a generalized generation algorithm for Shake and Bake that can accommodate the more general derivational structures used in HPSG, we then examine ways in which the efficiency of the translation process can be improved.

## 2.0  The Shake and Bake Algorithm

In S&B MT, the parsing of the source language sentence is performed with a unilingual grammar (and lexicon), as is the generation of the target language sentence. Instead of the parser producing an interlingual representation (to be used by a target language generator as is done in the *interlingua* approach), or a language specific synactico-semantic representation (to be subjected to language-pair specific transfer rules as is done in the *transfer* approach), the goal is to produce a bag of leaves (which are each *signs*, adopting the terminology used in UCG and HPSG) which are obtained from the derivational structure produced by the parser. Each of these leaves will have more information than its counterpart from the unilingual lexicon due to the unification and structure sharing that occurs as a result of a successful parse.

For example, consider a lexical entry for *love* which might be found in the lexicon of a lexical unification based formalism like UCG or HPSG. It will specify that this entry is a verb which subcategorizes for a subject and object with some as yet unconstrained semantic index or variable. After a successful parse, these semantic variables in the lexical entry for love will be unified or structure shared with the semantic indices found in the actual subject or object.

Given a bag of leaves obtained from the analysis of the source language sentence, a bilingual (or multilingual) lexicon is then used to obtain a corresponding bag of target language signs. Entries in the bilingual lexicon associate sets of target language signs with sets of source language signs. For example, the multilingual dictionary will match the English signs for the words *look* and *for*, with the single French sign for *cherche*. For the sake of translation, inflectional information can also be associated with separate signs. This mapping is not necessarily unique though. For a single source language bag, it may be possible to obtain several target language bags. By having a more constrained (and more complicated) bilingual lexicon, it is possible to constrain the number of target language bags and the size of the bags. However, a detailed discussion of the transfer phase is beyond the scope of this paper.

Finally, the target language bag is given to the S&B generation algorithm, which together with the unilingual grammar (and principles in the case of HPSG) for the target language is used to generate the sentence. For UCG, a variation of the shift-reduce parser is used to generate a sentence. Sets of two signs are selected from the bag, and then there is an attempt to apply a binary grammar rule to these signs. If the attempt is successful, the two original signs are removed and resulting sign is introduced into the bag. Although the generation algorithm is inefficient (in fact, the generation problem is NP-complete), it is possible to prune the generation search tree by utilizing additional constraints with the goal of obtaining acceptable results on realistic inputs [2]. Brew proposes the addition of a constraint module, which greatly reduces the search space but at the cost of the overhead associated with a constraint propagation algorithm.

### 3.0  Generalized Shake and Bake Generation

For HPSG, the generation algorithm can be slightly modified so that instead of combining two signs from the bag and attempting to apply a grammar rule, different sized sets of signs can be combined depending on the number of constituents allowed by the specific grammars and lexicons. This is very simple to implement, but with it comes a massive decrease in efficiency, since we now have to consider a huge number of possible sets of signs which might be subject to the application of a grammar rule. This is compounded by the fact that we need to consider all the different possible orderings of the selected signs when determining if a grammar rule can be applied. Furthermore, we need to incorporate the

```
%  sNb(+Stack, ?FinalSign, +BagIn, -BagOut)
%
sNb([Sign], Sign, [], []). %termination

sNb([First|Stack],Sign) --> % reduce
    {  difference(Stack, Others, Stack1),
       unordered_rule(Mother, [First|Others])},
    sNb([Mother|Stack1],Sign).

sNb(Stack,Sign) --> % shift
    [Next],
    sNb([Next|Stack],Sign).
```

Figure 1. Naïve Generation Algorithm

constraints associated with the different principles associated with an HPSG grammar.

A Prolog implementation of the generalized S&B generation algorithm is provided in Figure 1. A sign is *shifted* from the input bag to a 'stack', then a set of stack elements may be *reduced* by replacing them with the sign resulting from application of a grammar rule[1]. The algorithm is essentially identical to the one provided in [10], except that the relation **difference/3** is used to remove an arbitrary number of signs from the **Stack**, instead of the **delete/3** relation used by Whitelock to remove a single sign. The first clause states that the generation process has terminated when the bag is empty, and there is only one sign (the **Sign** corresponding to the complete sentence) contained in the stack. In the second and third clause, the DCG rule format is used, thus we do not need to explicitly show the **BagIn** and **BagOut** for each clause (the values for these variables are automatically threaded through the different clauses). The reduction clause nondeterministically selects *n-1* signs, **Others**, from the stack, where *n* is the maximum branching factor allowed in derivations by the grammar. For UCG *n=2*, while for HPSG *n* depends on the actual lexicon being used. An attempt is then made to reduce the **First** sign from the stack, together with the **Others** selected, by applying a grammar rule, **unordered_rule**. The processing of the constraints associated with the HPSG principles is done within **unordered_rule** as well. A successful reduction results in a new sign, **Mother**, replacing the **First** and **Others** in the stack. Finally, the third clause results in the **Next** sign from the bag being shifted to the top of the **Stack**. The entire generation process can be summarized graphically as shown in Figure 2.

To illustrate the naive generation algorithm for HPSG, let us consider the generation of the French sentence *Jean aime Marie* from the input bag consisting of the signs for *Jean, aime,* and *Marie*. In our example, we will use JEAN, AIME and MARIE to denote the signs. Let us assume, without loss of generality, that the bag is represented as the list of signs [AIME, MARIE, JEAN] (this is the actual bag produced by our simple transfer module). Initially, the stack will be empty, and the input bag will contain all three signs, as shown in Figure 3. First, AIME will be shifted onto the stack, and an attempt will then be made to reduce this single sign using the grammar rules. This attempt will fail, which will lead to the next sign, MARIE, being shifted onto the stack. Although an attempt to reduce just the top symbol of the stack, MARIE, will fail, a reduction of the top two signs will succeed resulting in a stack in which these two signs are replaced with the sign for AIME-MARIE (obtained via an HPSG grammar rule and the associated HPSG principles). Then, a final shift operation followed by a reduction is all that is needed to obtain a sign for the complete sentence as shown in Figure 3. Note that it would also be possible during backtracking to perform a shift (using the third clause of **sNb**) instead of a reduce when the stack contains [MARIE,AIME] as shown at the asterisk in the Figure. This choice will not lead to successful generation of a sentence in this example though.

---

1. As Whitelock notes, if this were a shift-reduce parser we would be shifting from the input string rather than a bag. Additionally, in a parser the reductions would be applied to only the top *n* elements of the stack, rather than the top element and some *n-1* other elements of the stack as is done in the generator. For this reason, it might be best to think of the stack used in the generator as a *reduction bag*, which has a distinguished first (top) element which must be included in the set of elements chosen for reduction. Aside from these differences, the shift reduce parsing and generation algorithms are essentially the same.

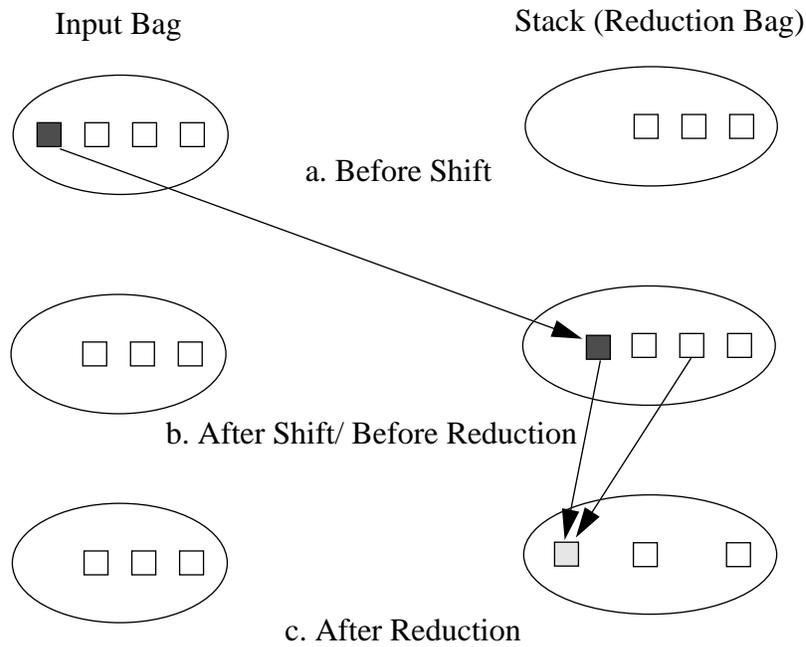

Figure 2. The Generation Process

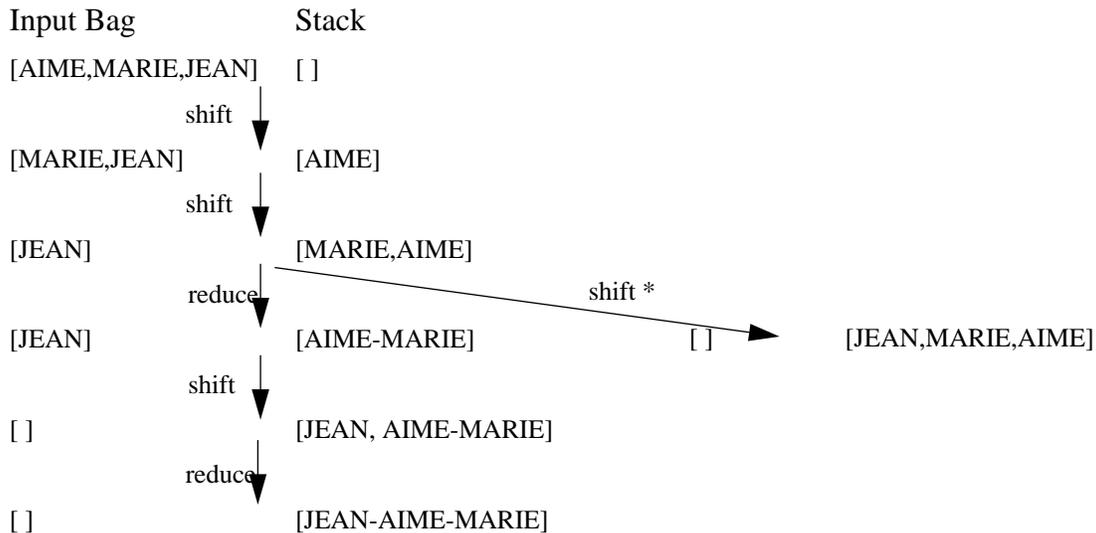

Figure 3. Example of Generation

### 4.0  Issues in Efficiency

There are three points in the translation process where we can address efficiency issues. The first is in the parsing of the sentence using the source language grammar, the second is in the generation of a target language bag from a source language bag using the multilingual lexicon, and the third is in the generation of a target language sentence from the target language bag using the target language grammar. In this paper, we will be focussing on improving the efficiency of the generation process, but we will suggest how some of techniques might be applied to the second stage as well.

As was noted above, S&B generation can be viewed simply as a variation of parsing, except that instead of having a sequence of words as input, one has a sequence of signs. As with a shift reduce parser, a shift reduce generator can find itself, due to backtracking, in a state that it had encountered in a previous attempt to reduce a collection of signs. For example, when the generator backtracks on the example in Figure 3 and finds itself in the state resulting from the shift operation marked with the asterisk, it will attempt to reduce the first element of the stack (which it would have already tried in the other branch in Figure 3) before trying to reduce it together with the other elements of the stack. We need to dramatically decrease the search space, without introducing too great of an overhead, if we want to obtain acceptable performance from an S&B generator.

A common strategy for improving the performance of search algorithms is to introduce a table to keep track of promising (or unpromising) states. This is commonly done in parsing, in the form of well-formed substring tables, or charts.[2] Such a technique is also commonly adopted in logic programming tasks, under the title of **memoizing** – storing the results of expensive computations so that if they are ever encountered again, their values can simply be retrieved rather than recalculated.

So, let us consider exactly where the efficiency of the above algorithm can be improved by keeping track of intermediate results. In examining the code provided above, or by having the Prolog system do a profile of its execution during generation, one finds that a great deal of effort is spent in the execution of the **unordered_rule** predicate. This predicate is used to see if a collection of signs from the stack can be combined according to any one of the grammar rules. Given a set of *n* signs as input arguments (the top sign of the stack and *n-1* others), the predicate tries all possible permutations of these signs to see if they can be valid daughters (constituents) according to any of the grammar rules. In terms of the example that we introduced in Figure 3, whenever the stack contains one element, **unordered_rule** will be called once. When it contains two elements, **unordered_rule** will be called twice (once for the top element of the stack, and once for the top element and the other element), and when the stack contains three elements there will be four calls. For a stack of size *s* and with a maximum branching factor of *n*, the number of calls to **unordered_rule** during a reduction is determined according to the equation shown in (1), a summation of the ways that *i-1* elements can be chosen from a set of *s-1* elements. Since

$$(1) \quad \sum_{i=1}^{min(n,s)} \binom{s-1}{i-1}$$

most HPSG grammars have four or five grammar rules (or schemas), and have the value of *n* being either three or four, it suggests that this predicate is an appropriate place to introduce some method to avoid the duplication of work. But we need to ensure that the overhead does not compete with the benefits.

---

2. An introduction to how these techniques are used in Prolog can be found in [4], while [7] shows how a similar technique can be used to accumulate failure results during parsing.

The way that the generation problem is structured, most of the time an attempted application of **unordered_rule** will fail. Therefore, whenever we apply this predicate to a set of signs and fail to produce a resulting sign, we will keep track of this failure. So if in the future there is an attempt to apply **unordered_rule** to this same set of signs given in any order, we can immediately signal failure. In the few cases where the application of **unordered_rule** does succeed, we can store the resulting structure or structures (there need not be only one way in which the signs can be successfully combined) for future use.

This process can be incorporated into the generation algorithm by associating a unique integer (which we will call a tag) with each sign in the bag (initial signs and those generated during the generation process). Then, the results obtained by calling **unordered_rule** on a set of signs can be stored and retrieved based on the tags associated with the signs. Instead of indexing though on a list of tags, it is more efficient (as we will see in the next section) to calculate a unique integer index[3] for the list of tags according to the formula shown in (2) where each *t* is a tag. The approach used to implement this function is shown in Figure 4.

$$(2) \quad index(\langle t_1, t_2, ..., t_n \rangle) = \sum_{i=1}^{n} 2^{t_1 - 1}$$

Note that exponentiation is obtained by a *shift left* operation being performed on integers.

We then only need to provide a definition for **unordered_rule**, which will access a stored result if one is available, or calculate and store the result if one is not available. The definition for this predicate is supplied in Appendix A.

---

3. Since we are using integers, the technique that we are using is limited by the size of the maximum integer supported in the Prolog implementation. In Sicstus Prolog, the maximum integer is 2^2147483616, thus number of signs used in the generation process is essentially unlimited. Preliminary results suggest that the **calculate/2** relation is also slightly more efficient than using the built-in Sicstus predicate **term_hash/3**. However, our point here is not to compare different hashing functions, but to show that it is important to index a clause on a simple key for efficient retrieval. We are essentially encoding subsets of signs as bitstrings which we use to access our table of stored results.

**5.0  Preliminary Experiments**

In order to test the HPSG translation algorithm and examine the behavior of the generation algorithm, we made some minor modifications to the HPSG-PL system [6]. It was necessary to modify the parser so that it would produce a bag of lexical signs (performed by Kodric and Popowich), plus it was necessary to write a transfer component and implement the generation algorithm described in this report. A small French grammar, equivalent to a subset of the small English grammar provided with the HPSG-PL system, was produced, and a simple bilingual lexicon used in the creation of a target language bag from a source language bag was also developed. The various lexicons are supplied in Appendix B. Separate Prolog modules were created for the use of each language. The HPSG-PL lexical compiler was used to convert each lexicon into an internal representation for use by the generators and parsers. The generators and parsers for each language differed only in terms of the HPSG language specific principles that they used.

In order to evaluate the success of the storage of previous results, calls to **unordered_rule** were monitored to see if they involved the retrieval of a previously stored result (a *hit*), or the calculation and storage of a new result (a *miss*). These calculations were performed for a three word sentence *John loves Mary,* a six word sentence *Kim gives the cookie to Mary* and a nine word sentence *Mary gives the good cat to the small girl*. The second and third sentences had analyses in which branching factors of one, two and three were present. The translator used a very simple bilingual lexicon which generated one target language bag for the first sentence, two for the second, and sixteen for the third[4]. In each case though, only one of the target bags could be used to generate a grammatical sentence. By introducing more information into the bilingual lexicon, it would be possible to highly constrain the number of target bags generated, as is in done [10]. Nevertheless, even the bags

```
calculate(TagList, Value) :-
   calc_aux(TagList, 0, Value).

calc_aux([], Value, Value).

calc_aux([I|TagList], In, Out) :-
   Value is (1 << (I-1)) + In,
   calc_aux(TagList, Value, Out).
```

Figure 4. Indexing of Tags

---

4. In the bilingual lexicon, there is one entry pairing the English definite article *the* with the masculine French definite article *le*, and another entry pairing it with the feminine French definite article *la*. There are similar entries for masculine and feminine versions of the adjectives. Since the second sentence contained one determiner, there were two possible target bags (one with the masculine determiner and one with the feminine), and since the third sentence contained four determiners or adjectives, there were sixteen possible bags. It was up to the French language grammar to determine which translation is appropriate. This disambiguation could also be done during the creation of the bag of French signs, before the generation process commences. However, in this paper we are concerned with the generation process itself, and thus have not yet constructed the more elaborate bilingual lexicon needed for this technique.

which did not result in a grammatical sentence were useful for examining the behavior of the generator.

The performance of the generator for cases where it was possible to generate a grammatical sentence from the input bag of signs provided is shown in Table 1. The table first shows the size of the input bag and the number of calls to **unordered_rule.** Then the times required (with and without the use of the technique for storing previously calculated results) to generate the first sentence from the input bag are shown, followed by the total time (which includes the time for the first sentence) spent trying to generate all possible sentences. The table also summarizes the difference in performance when formula (2) is used to index the stored results as opposed to a list of tags associated with the constituent signs. Finally, the table shows the number of hits, followed by the hit ratio; the number of times that a stored result is used, and how this compares to the total number of calls to **unordered_rule**. All of the tests were run under Sicstus Prolog 2.1 #9 on a Sun SparcStation 20.

Table 1: Generation Results

| | size of bag | no. of calls | without result storage | | with storage of results | | | | hits | hit ratio |
|---|---|---|---|---|---|---|---|---|---|---|
| | | | | | integer index | | list of tags | | | |
| | | | 1st (sec) | tot (sec) | 1st (sec) | tot (sec) | 1st (sec) | tot (sec) | | |
| 1 | 3 | 11 | 0.02 | 0.04 | 0.02 | 0.04 | 0.02 | 0.04 | 1 | 0.09 |
| 2 | 6 | 220 | 0.15 | 0.94 | 0.16 | 0.69 | 0.18 | 0.95 | 90 | 0.41 |
| 3 | 9 | 2883 | 0.92 | 13.67 | 0.70 | 5.51 | 1.02 | 18.16 | 2009 | 0.70 |

### 6.0 Conclusions

Our results are very preliminary, and additional results need to be obtained with expanded grammars and lexicons. It would also be instructive to run experiments on grammars with varying number of rules and varying branching factors. However, the high hit ratio on longer sentences together with the pronounced decrease in total time when using the storage looks promising. Although the use of the store does require additional processing time, it is insignificant compared to the amount of effort it saves. Our results also suggest that the technique may be practical for shorter sentences that can be generated from bags of less than a dozen constituents, but the practicality for long sentences has yet to be illustrated.

Although we applied our storage technique only to the generation process, it might also be used in cases where the lexical transfer process results in more than one possible bag to which the generation algorithm needs to be applied. Consider the case where there are two possible bags, each having some signs in common. Results obtained during calculations involving the signs from the first bag might also be useful during calculations involving the second (and even later) bags.

**Acknowledgements**

I would like to thank Dan Fass, Sandi Kodric, Olivier Laurens, Paul McFetridge, Carl Vogel and the anonymous referees for their comments and suggestions.This research was supported by a Research Grant from the Natural Sciences and Engineering Research Council of Canada.

## Appendix A  Predicate Definitions

```
% unordered_rule(-TMother, +TDtrs)
%
% given a list of daughter signs (which are tagged with integers),
% first see if we have encountered this list before. If so, then
% use the check_found/3 to either get the value, or determine if the
% previous attempt failed. Otherwise, (final clause) we will have to
% actually try to apply a rule, and store the result we obtain.

unordered_rule(TMother, TDtrs) :-
 get_tag_list(TDtrs, TagList, Dtrs),
 unordered_rule(TagList, TMother, Dtrs).

unordered_rule(TagList, TMother, Dtrs) :-
 check_found(TagList,TMother,Success_or_Fail) -> % we've done it before
 inc(hit), % increment our counter
 call(Success_or_Fail); % should we succeed or fail?

 % Since, table lookup fails...
 % We must consider all possible permutations of the daughters.

 not_found(TagList), % assume things won't succeed
 inc(miss),
 rule(_RuleNumber, Mother, Constraints),
 permutation(Dtrs, [Head|Args]),
 get_dtrs(Mother, Head, Args),
 process(Constraints), % process the HPSG constraints (& principles)
 tag(Mother, TMother), % assign a number to the new sign
 is_found(TagList, TMother). % if we actually do succeed

% Our chart makes use of the dynamic predicate dtrs_lookup/3

check_found(TagList, TMother, Success_or_Fail) :-
 calculate(TagList, Value),
 dtrs_lookup(Value, TMother, Success_or_Fail).

not_found(TagList) :-
 calculate(TagList, Value),
 assert(dtrs_lookup(Value,_,fail)).

is_found(TagList,TMother) :-
 calculate(TagList, Value),
 retract(dtrs_lookup(Value,_,fail)),
 assert(dtrs_lookup(Value,TMother,true)).
```

## Appendix B  Lexicons

1. English Lexicon

```
entry @adject(good).
entry @adject(small).

entry @cn(cat).
entry @cn(cookie).
entry @cn(table).
entry @cn(man).
entry @cn(woman).
entry @cn(girl).

entry @det(the).

entry @pn(john).
entry @pn(kim).
entry @pn(mary).

entry @preposition(on).
entry @preposition(to).

entry @intransv(walks,@fin).
entry @intransv(sleeps,@fin).

entry @transv(eats,@fin).
entry @transv(loves,@fin).

entry @ditransv(gives,@fin).
```

2. French Lexicon

```
entry @adject(bon,masc).
entry @adject(bonne,fem).
entry @adject(petit,masc).
entry @adject(petite,fem).

entry @cn(chat,masc).
entry @cn(biscuit,masc).
entry @cn(table,fem).
entry @cn(homme,masc).
entry @cn(dame,fem).
entry @cn(fille,fem).

entry @art(le,masc).
entry @art(la,fem).

entry @pn(jean).
entry @pn(kim).
entry @pn(marie).

entry @preposition(sur).
entry @preposition(a).

entry @intransv(marche,@fin).
entry @intransv(dort,@fin).

entry @transv(mange,@fin)..
entry @transv(aime,@fin).

entry @ditransv(donne,@fin).
```

## 3. Bilingual Lexicon

```
john entry @semindex(X) <==> jean entry @semindex(X).
mary entry @semindex(X) <==> marie entry @semindex(X).
kim entry @semindex(X) <==> kim entry @semindex(X).

eats entry @cont_args(Args) <==> mange entry @cont_args(Args).
gives entry @cont_args(Args) <==> donne entry @cont_args(Args).
loves entry @cont_args(Args) <==> aime entry @cont_args(Args).
sleeps entry @cont_args(Args) <==> dort entry @cont_args(Args).
walks entry @cont_args(Args) <==> marche entry @cont_args(Args).

cat entry @semindex(Args) <==> chat entry @semindex(Args).
cookie entry @semindex(Args) <==> biscuit entry @semindex(Args).
table entry @semindex(Args) <==> table entry @semindex(Args).
girl entry @semindex(Args) <==> fille entry @semindex(Args).

the entry @semindex(X) <==> le entry @semindex(X).
the entry @semindex(X) <==> la entry @semindex(X).

good entry @semindex(X) <==> bon entry @semindex(X).
good entry @semindex(X) <==> bonne entry @semindex(X).

small entry @semindex(X) <==> petit entry @semindex(X).
small entry @semindex(X) <==> petite entry @semindex(X).

to entry @semindex(X) & @prep
 <==> a entry @semindex(X) & @prep.
```